\documentclass[conference]{IEEEtran}
\IEEEoverridecommandlockouts

\usepackage{subcaption}

\usepackage{colortbl}
\usepackage{xcolor}

\usepackage{cite}
\usepackage{amsmath,amssymb,amsfonts}
\usepackage{algorithmic}
\usepackage{graphicx}
\usepackage{textcomp}
\usepackage{xcolor}
\def\BibTeX{{\rm B\kern-.05em{\sc i\kern-.025em b}\kern-.08em
    T\kern-.1667em\lower.7ex\hbox{E}\kern-.125emX}}
\begin{document}

\title{Renormalized Graph Representations for Node Classification
}



\author{
    \IEEEauthorblockN{Francesco Caso}
    \IEEEauthorblockA{
        DIAG\\
        University of Rome, La Sapienza\\
        Email: francesco.caso@uniroma1.it
    }
    \and
    \IEEEauthorblockN{Giovanni Trappolini}
    \IEEEauthorblockA{
        DIAG\\
        University of Rome, La Sapienza\\
        Email: giovanni.trappolini@uniroma1.it
    }
    \and
    \IEEEauthorblockN{Andrea Bacciu}
    \IEEEauthorblockA{
        DIAG\\
        University of Rome, La Sapienza\textsuperscript{*}\thanks{\textsuperscript{*}Work done prior to joining Amazon.}\\
        Email: andrea.bacciu@uniroma1.it
    }
    \and
    \IEEEauthorblockN{Pietro Liò}
    \IEEEauthorblockA{
        Department of Computer Science and Technology\\
        University of Cambridge\\
        Email: pl219@cam.ac.uk
    }
    \and
    \IEEEauthorblockN{}
    \IEEEauthorblockA{
        \quad
    }
    \and
    \IEEEauthorblockN{\qquad \qquad \qquad Fabrizio Silvestri}
    \IEEEauthorblockA{
        \qquad \qquad \qquad DIAG\\
        \qquad \qquad \qquad University of Rome, La Sapienza\\
        \qquad \qquad \qquad Email: fabrizio.silvestri@uniroma1.it
    }
}

\maketitle

\begin{abstract}
Graph neural networks process information on graphs represented at a given resolution scale. We analyze the effect of using different coarse-grained graph resolutions, obtained through the Laplacian renormalization group theory, on node classification tasks. At the theory's core is grouping nodes connected by significant information flow at a given time scale. Representations of the graph at different scales encode interaction information at different ranges. We specifically experiment using representations at the characteristic scale of the graph's mesoscopic structures. We provide the models with the original graph and the graph represented at the characteristic resolution scale and compare them to models that can only access the original graph. Our results showed that models with access to both the original graph and the characteristic scale graph can achieve statistically significant improvements in test accuracy.\\ Code and supplementary material are available at the following link: https://anonymous.4open.science/r/RGR-4906/README.md
\end{abstract}


\section{Introduction}
Graphs are an abstract representation of elements, often referred to as nodes, and their pairwise interactions, or edges. Edges connect a source node to a destination node, and the set of destinations of a given source is called its neighborhood. Considered individually, neighborhoods encode the information of local interactions, but from the intersections of neighborhoods of different nodes, the graph develops its own geometry. 
The graph's geometry manifests through structures larger than edges and provides insights into implicit interactions between nodes that are not directly linked by an edge but can be reached through a path—a sequence of edges where the destination of the previous edge is the source of the next edge. Various mesoscopic graph structures—larger than a single edge and smaller than the entire graph—encode interactions with increasingly wider ranges \cite{bullmore2012economy,lyu2024learning}.
It is possible to consider mesoscopic structures as local interactions by rescaling the graph—i.e. creating a new graph representation where the nodes are groups of the original nodes, and the edges represent long-range interactions \cite{villegas2023laplacian}.
The scaling transformations of a physical system and the changes in the mathematical models that represent it are formalized in the Renormalization Group (RG) theory \cite{mussardo2020statistical}. 
A RG for complex networks is still an open problem. Some remarkable initial attempts proposed to renormalize the network by embedding it in an underlying hidden metric space \cite{garcia2018multiscale}. By defining a metric, the clusters of nearby nodes to be grouped are determined. However, previous attempts struggle both in real-world cases and in known theoretical cases \cite{villegas2023laplacian}. For example, in real networks, they often cannot be applied iteratively due to their small-world property, which intertwines different scales of the network. In theoretical cases, it has been observed that these renormalization procedures fail to preserve the average degree of scale-free networks, such as Barabási-Albert networks, in coarse-grained replicas. A different renormalization approach leverages diffusion distances \cite{villegas2022laplacian}. In this context, the Laplacian Renormalization Group (LRG) \cite{villegas2023laplacian} has been shown to preserve properties across scales in scale-free networks and can be applied iteratively to real small-world networks. At the core of this approach is the use of diffusion to identify nodes belonging to the same cluster. However, this approach differs from previous ones by leveraging more recent tools from graph information theory \cite{de2016spectral}, which allow clusters to be interpreted in terms of their roles in dynamical processes defined on the graph \cite{de2014identifying, viamontes2011compression, rosvall2008maps}. In particular, by analyzing the variation in the speed of information propagation, mesostructures can be identified. These are regions where information quickly becomes uniform during diffusion, indicating nodes that are strongly connected at a mesoscale level.
In the realm of deep learning, Graph Neural Networks (GNNs) are models designed to process information on graphs. Diffusion on graphs has often been linked to information processing on graphs and to GNNs \cite{shuman2013emerging, gasteiger2019diffusion, chamberlain2021grand}. Reference \cite{gasteiger2019diffusion} used diffusion as a preprocessing step for graphs that GNNs later process. In contrast, \cite{chamberlain2021grand} drew an analogy between the layers of GNNs and graph diffusion, which becomes an equivalence for certain specific choices of layer parameters. At the same time in graph learning, the graph is often processed at its given resolution. This means the deep learning models will have explicit access to local edge interactions and implicit access to structures emerging from iterative message passing. In principle, long-range structures can always be reconstructed through successive applications of local models. However, to simplify the network's learning and generalization, long-range information can be explicitly introduced. Following the connection between GNNs and diffusion, in this work, we study the use of incorporating long-range information by providing the model not only with the original graph representation but also with a coarse-grained version obtained through the LRG in node classification tasks. In general, analyzing a system on the correct resolution scale is crucial for finding an effective representation \cite{lepage1989renormalization, fisher1974renormalization, delamotte2004hint} and the mesoscopic structures of graph topology are often underutilized in node-level tasks. When they are employed, they are rarely based directly on principles and theories \cite{luzhnica2019clique, lipov2021multiscale}. Instead, they typically rely on tunable parameters \cite{abu2020n}. To the best of our knowledge, we are the first to propose a method for utilizing the mesoscopic structures of a graph based on the LRG theory. 

Assuming that dividing the model into graph processing at different scales can be a useful inductive bias, we contribute to the study of node classification by:  
\begin{itemize}
    \item Defining a framework that allows the model to consider multiple scales in parallel and then combine the results. The framework consists of a method to represent a graph at a certain scale using the LRG, followed by concatenating the information processed separately at different scales.
    \item Demonstrating the experimental usefulness of this framework. 
    \item Showing that the study of spectral entropy can provide relevant scales a priori, before any training procedure.
\end{itemize}
In the following, we define how we preprocess the graph to align it with scale, effectively using the characteristic scales identified by \cite{villegas2022laplacian}. Subsequently, we present the framework of models that analyze multiple scales in parallel. After a description of the datasets and the experimental setup, we present results demonstrating the utility of attending to multiple scales of the graph.   

%

\section{Graph Rewiring}



After defining the notation used throughout the paper, we summarize the underlying ideas for identifying the characteristic scale and we then present the preprocessing method that aligns the graph with a given characteristic scale, through a rewiring of the graph.\\

\paragraph{Notation} We define a graph to be $\mathcal{G} = (\mathcal{V}, \mathcal{E})$ where $\mathcal{V}$ is the set of nodes of the graph, and $\mathcal{E} \subseteq \mathcal{V} \times \mathcal{V}$ is the set of edges. We will consider only unweighted graphs so the set of edges can also be represented through an adjacency matrix $A \in \mathcal{R}^{|\mathcal{V}|\times |\mathcal{V}|}$, whose elements are 
\begin{equation}
    A_{ij} = \begin{cases}
    1 & (i,j) \in \mathcal{E}\\
    0 & (i,j) \notin \mathcal{E}

    \end{cases} \quad . 
\end{equation}
We consider the Laplacian matrix defined as 
\begin{equation}
    L_{ij} = [(\delta_{ij}\sum_k A_{ik})-A_{ij}]
\end{equation}
where $\delta_{ij}$ is the Kronecker delta function.
LRG theory is limited to undirected graphs with a single connected component \cite{villegas2023laplacian}.\\

\begin{figure}[h]
    \centering
    \includegraphics[width=\linewidth]{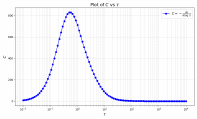}
    \caption{The heat capacity, defined as the derivative of spectral entropy with respect to the logarithm of time, is plotted as a function of time. The peak in this plot identifies the characteristic time scale. The values are related to the Cora graph.}
    \label{fig:specific_heat}
\end{figure}

\paragraph{Characteristic scales} The way GNNs process information has been compared to information diffusion on graphs \cite{chamberlain2021grand}. Typically, information diffusion on a graph exhibits two (or more) qualitatively different behaviors, corresponding to two (or more) phases, as observed by \cite{villegas2022laplacian}. To observe the phase transition, one can consider the propagator of the diffusion
\begin{equation}
    \rho(\tau) = \frac{e^{-\tau L}}{Tr(e^{-\tau L})} \quad ,
\end{equation}
such that the vector representing each node at time $\tau$ is given by $v(\tau)=\rho(\tau) v(0)$. Subsequently, the von Neumann entropy of the propagator
\begin{equation}
    S[\rho(\tau)] = - \frac{1}{\log(N)} \sum_{i=1}^N \mu_i(\tau)\log \mu_i(\tau) \quad ,
\end{equation} 
which is a function of its eigenvalues $\mu_i(\tau)$, is analyzed, as in \cite{de2016spectral}. The phase transition occurs when the derivative of the entropy with respect to the logarithm of time 
\begin{equation}
    C = - \frac{dS}{d(\log \tau)} \quad ,
\end{equation}
shows a peak, as shown in Figure \ref{fig:specific_heat}. This behavior signifies that, shortly after this point, there is a marked deceleration in information diffusion that distinguishes strongly connected regions, i.e. mesoscopic structures, from the rest of the network. 
\paragraph{Renormalized graph representations} To preprocess the graph so that it represents one of the previously identified characteristic scales, we follow a renormalization procedure inspired by \cite{villegas2023laplacian}. First, we calculate the diffusion propagator, which is a non-trivial function of the Laplacian and represents the information exchanged between two nodes through diffusion. Next, we define groups of nodes that will represent single nodes in the system at a lower resolution, referred to as "macro-nodes". 

\begin{figure*}[ht!] 
    \centering
    \begin{subfigure}[b]{0.23\textwidth} 
        \centering
        \includegraphics[width=\textwidth]{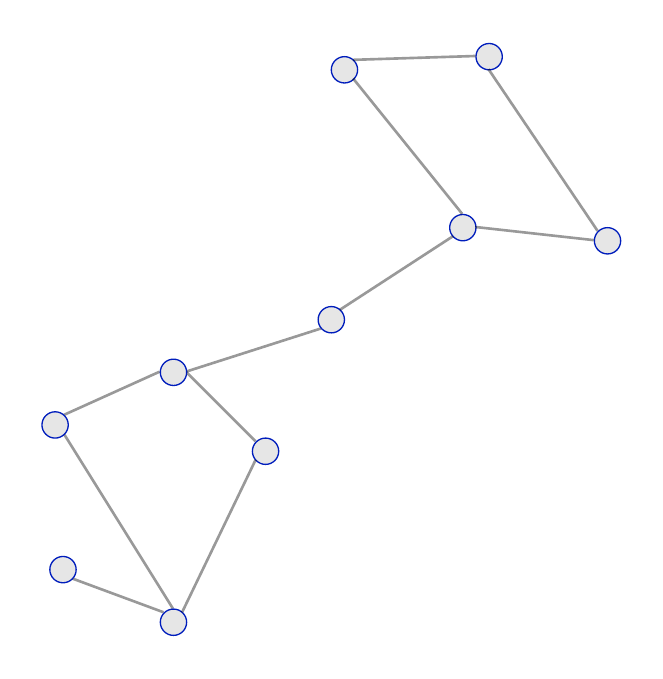}
        \caption{Original}
        \label{fig:sub1}
    \end{subfigure}
    \hfill
    \begin{subfigure}[b]{0.23\textwidth}
        \centering
        \includegraphics[width=\textwidth]{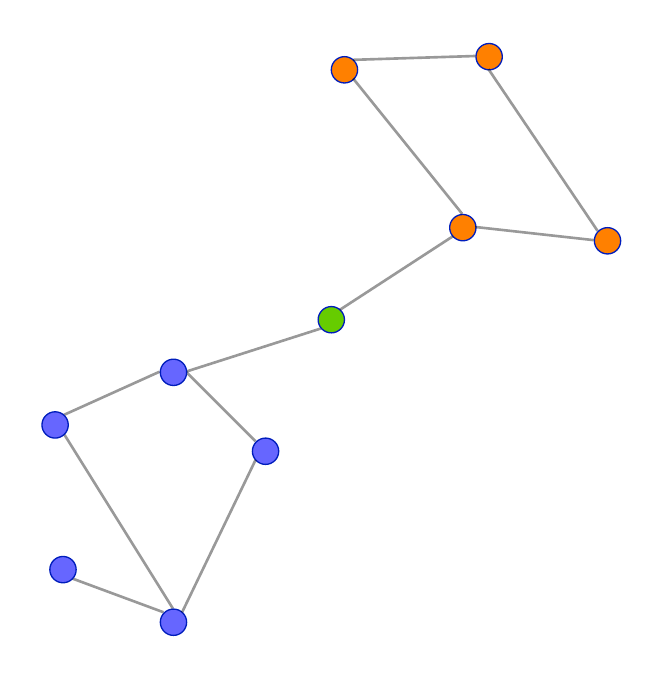}
        \caption{Clustered}
        \label{fig:sub2}
    \end{subfigure}
    \hfill
    \begin{subfigure}[b]{0.23\textwidth}
        \centering
        \includegraphics[width=\textwidth]{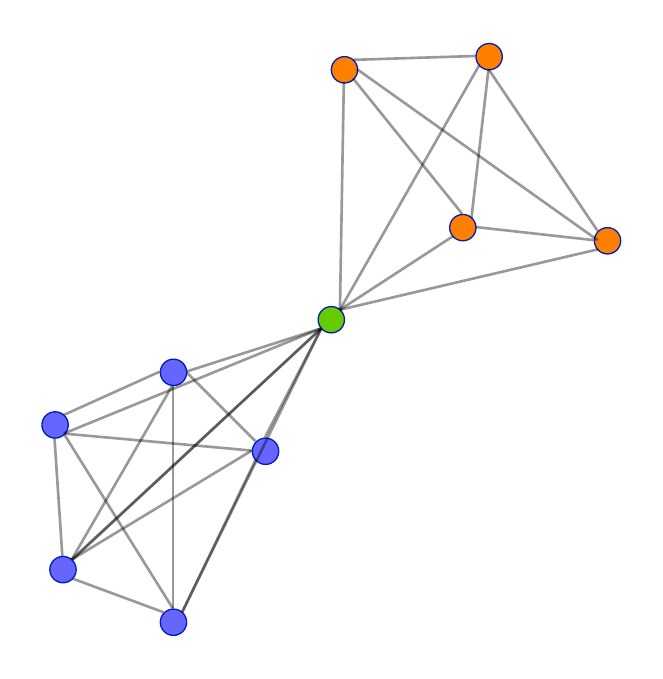}
        \caption{Rewired}
        \label{fig:sub3}
    \end{subfigure}
    \hfill
    \begin{subfigure}[b]{0.23\textwidth}
        \centering
        \includegraphics[width=\textwidth]{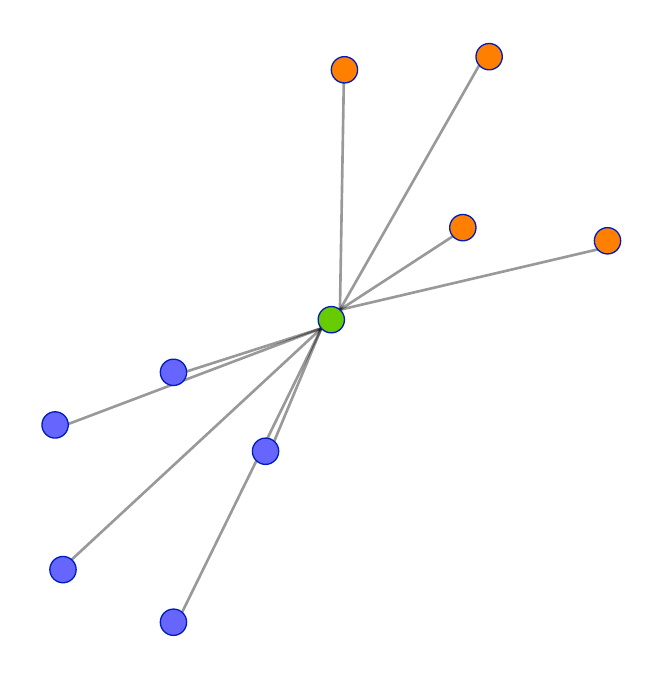} 
        \caption{Sparsified}
        \label{fig:sub4}
    \end{subfigure}

    \caption{A visualization of the proposed rewiring procedure. The nodes of the original graph (left) are clustered in three macro-nodes (second from left). The graph is rewired in such a way that nodes in the same macro-node share the same incoming and outgoing edges (third from left). Finally, the links between nodes of the same macro-node are removed to sparsify the graph (right).}
    \label{fig:rewiring}
\end{figure*}

To determine which nodes belong to the same macro-node, we examine the mutual information, represented by the elements of the propagator $\rho_{ij}$, between each pair of nodes. If the mutual information $\rho_{ij}$ is greater than the self-information of one of the two nodes $\min\{\rho_{ii}, \rho_{jj}\}$, they should belong to the same macro-node. Once the macro-nodes are defined, we modify the graph so that nodes within the same macro-node share the same neighborhoods. We finally remove the links between nodes within the same macro-node to sparsify the graph. 
Figure \ref{fig:rewiring} provides a sketch to help understand the procedure: using the propagator matrix, we check whether two nodes should belong to the same macro-node. This uniquely defines macro-nodes, which are represented in the figure by different colors. Nodes belonging to the same macro-node must share the same neighborhood, which is formed by merging the neighborhoods of all nodes in the macro-node. Finally, we remove the links between nodes within the same macro-node to sparsify the graph. 

In this way, we are able to first identify one or more characteristic scales of the graph, representing mesoscopic structures of its topology. Then, we modify the graph's links so that it represents the same complex system, but at one of the characteristic resolution scales. Ultimately, for each complex system, we obtain a set of graphs—one for each characteristic scale—comprising the original graph and the versions aligned with each of the characteristic scales.\\

\paragraph{Attending more representations}
Given a set of graphs representing the same system, each aligned with a different scale, we present here a framework for a generic model that can leverage all these scales. The inspiration comes from RG theory, which approximates complex nonlinear relationships through a weighted sum of simpler terms, each associated with a different scale \cite{delamotte2004hint}.

\begin{figure}[h]
    \centering
    \includegraphics[width=0.6\linewidth]{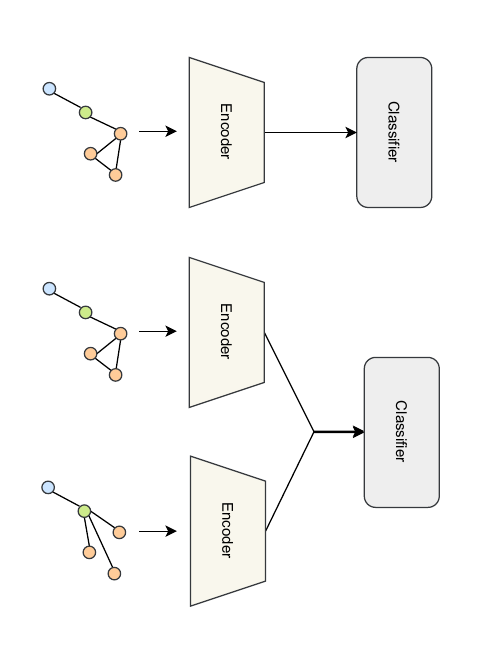}
    \caption{Representations of the tested models: one that focuses on a single scale (i.e., a typical GNN encoder + classifier) is shown at the top, while a model that considers two graph scales is shown at the bottom. Note that the two input graphs in the model at the bottom represent the same phenomenon, but viewed at different scales.}
    \label{fig:architecture}
\end{figure}

The test framework we defined to evaluate the utility of different scales in graph learning is based on classifying the nodes of a graph viewed at multiple scales in parallel. The main feature of the evaluated model is its ability to process multiple graphs simultaneously, each representing a different scale. The graph encoder class consists of one or more layers of the same type of GNNs. Each graph encoder processes a graph that is aligned with only one characteristic scale. The outputs of each graph encoder are then concatenated and passed to a classifier to produce the final result, as shown in Figure \ref{fig:architecture}.

Now that we have defined how to identify the characteristic scales and modify the graphs to align with these scales, as well as specified the framework of models that can utilize different scales of the graph, we present the details of the experiments conducted to verify the utility of the various graph scales in node classification tasks.

\section{Experiments}
We present the experiments we conducted to address the following questions:
\begin{description}
    \item[RQ1:]Is it beneficial to observe multiple scales of a graph for performing a node classification task?
    \item[RQ2:]Can we systematically identify the optimal scales using spectral entropy?
\end{description}
We also present and comment the results.

\subsection{Datasets}
We conducted node classification experiments on the following datasets: \textit{Citeseer} \cite{giles1998citeseer}, \textit{Cora} \cite{mccallum2000automating}, \textit{PubMed} \cite{namata2012query}, \textit{Europe} \cite{ribeiro2017struc2vec}, \textit{Amazon Photo} \cite{mcauley2015image} and \textit{Amazon Computers} \cite{shchur2018pitfalls}. The first 3 are citation networks; \textit{Europe} is an Air-traffic network; the last 2 are co-purchased products networks. The description of all datasets is provided in Table \ref{tab:datasets}. 

\begin{table}[h]
\caption{Datasets' description} \label{datasets-table}
\begin{center}
\begin{tabular}{lllll}
\textbf{Name}  &\textbf{edges} &\textbf{nodes} &\textbf{features} &\textbf{classes}\\
\hline \\
Citeseer             &4732   &3327  &3703   &6 \\
Cora         &5429   &2708  &1433   &7 \\
Europe & 5995 & 399 & 399 & 4\\
PubMed             &44338  &19717   &500    &3 \\
Photo   &238162 &7650   &745    &8\\
Computers   &491722 &13752  &767    &10   \\

\label{tab:datasets}
\end{tabular}
\end{center}
\end{table}

\subsection{Model}
During our testing procedure, we used two-layer graph encoders, such as GCN or GAT, as the basic building blocks of the model. The entire model is constructed by composing multiple graph encoders of the same type in parallel. Each graph encoder can receive a different graph as input, for example, one that has been preprocessed to represent a different scale. The information from the various encoders is then concatenated and fed into the final classifier, in our case a linear layer.

For each model that processed \(n\) different scales simultaneously, we also evaluated a baseline model with the same number of encoders in parallel, but all observing only the original graph. This allowed us to separate the contribution of the increased number of parameters and the larger space for hidden representation from the benefit of using multiple graph scales. We also compared these "multi-models" with "single-models" consisting of a single graph encoder that processes the original graph or a renormalized representation of it.

To train our model, we do not perform hyperparameter tuning, and we opt instead for standard values used in the literature.
We use the Categorical Cross-Entropy, Adam Optimizer \cite{kingma2014adam}, with a standard learning rate of 0.0001, over 1000 epochs, using the highest accuracy achieved on the validation set as our checkpointing strategy.
We perform our experiments on a workstation equipped with an Intel Core i9-10940X (14-core CPU running at 3.3GHz) and 256GB of RAM, and a single Nvidia RTX A6000 with 48GB of VRAM.

\subsection{Methods}
The presented model framework was tested using GAT or GCN based graph encoders and evaluated on the datasets discussed. Our research questions focus on the potential usefulness of graph representations at different scales compared to the original one. Therefore, we compared models that are as similar and not too complex as possible to ensure reliable controls. In particular, we compared models that could see only one scale—either the original/base scale or the renormalized characteristic scale—as well as models that could see multiple graphs, either all in the original representation or in the characteristic scale representations. The nodes of the graphs were masked into a test set and a train set with a single split. Graphs that were not undirected initially were made undirected, and only the largest connected component was considered. For both the original and renormalized graph representations, during training, information was propagated only within the subgraph consisting of the training nodes. The experiments were repeated across 10 different seeds, and the scores for individual test nodes were recorded, with a score of one assigned when the predicted class matched the true class, and zero otherwise.

\subsection{Results}
With the previously presented procedure, we aimed to answer \textbf{RQ1}: is it beneficial to observe multiple scales of a graph for performing a node classification task? Table \ref{tab:comparison2} displays the average scores, analogous to the test accuracy, along with their standard deviations. In red, the highest average values are highlighted, and in orange, the second-highest values. More importantly, for each model, Wilcoxon tests \cite{wilcoxon1992individual} were performed on the list of scores for all individual nodes obtained from the 10 experiments. A plus sign is placed next to the results of the model using renormalized representations if the test was statistically significant at the 5\% level, with the alternative hypothesis that the model with access to multiple scales scored more often than the model processing only the original graph (possibly repeated). A minus sign is placed when the test with the opposite alternative hypothesis was statistically significant. If the model using renormalized representations has no signs next to it, it means it was statistically equivalent to the analogous model that does not use renormalized representations.

\begin{table*}[ht]
\centering
\definecolor{redtext}{RGB}{200, 0, 0} 
\definecolor{orangetext}{RGB}{255, 150, 0}
\begin{tabular}{c c c c c c c} 
\hline
 \textbf{Model} & \textbf{Citeseer} & \textbf{Cora} & \textbf{Europe} & \textbf{Pubmed}  & \textbf{Photo} & \textbf{Computers} \\ 
 \hline
 \multicolumn{7}{l}{\textbf{Base}} \\
 \hline
 \multicolumn{7}{l}{\textit{Single}} \\
 $\text{GAT}_{\text{SB}}$ & $65.6 {\tiny\pm2.0}$ & $72.6 {\tiny\pm1.7}$ & $34.2 {\tiny\pm5.7}$ & $71.1 {\tiny\pm1.5}$ & $95.9 {\tiny\pm0.2}$ & $88.0 {\tiny\pm1.5}$ \\ 
 $\text{GCN}_{\text{SB}}$ & $65.8 {\tiny\pm1.6}$ & $72.0 {\tiny\pm0.7}$ & $35.1 {\tiny\pm5.3}$ & $\textcolor{orangetext}{73.9 {\tiny\pm1.5}}$ & $96.0 {\tiny\pm0.2}$ & $\textcolor{orangetext}{92.0 {\tiny\pm0.3}}$ \\ 
 \hline
 \multicolumn{7}{l}{\textit{Multi}} \\
 $\text{GAT}_{\text{MB}}$ & $66.3 {\tiny\pm2.4}$ & $73.6 {\tiny\pm1.4}$ & $30.8 {\tiny\pm4.6}$ & $71.9 {\tiny\pm1.2}$  & $96.1 {\tiny\pm0.3}$ & $91.5 {\tiny\pm0.4}$ \\
 $\text{GCN}_{\text{MB}}$ & $65.7 {\tiny\pm2.7}$ & $72.4 {\tiny\pm1.1}$ & $\textcolor{orangetext}{35.8 {\tiny\pm5.1}}$ & $\textcolor{orangetext}{73.9 {\tiny\pm1.0}}$ & $\textcolor{orangetext}{96.2 {\tiny\pm0.2}}$ & $\textcolor{redtext}{92.6 {\tiny\pm0.2}}$ \\
 \hline
 \multicolumn{7}{l}{\textbf{Renormalized}} \\
 \hline
 \multicolumn{7}{l}{\textit{Single}} \\
 $\text{GAT}_{\text{SR}}$ & $67.5 {\tiny\pm1.7}^{+}$ & $74.2 {\tiny\pm0.9}^{+}$ & $30.1 {\tiny\pm7.7}^{-}$ & $72.2 {\tiny\pm1.1}^{+}$ & $92.1 {\tiny\pm0.7}^{-}$ & $73.0 {\tiny\pm4.8}^{-}$ \\
 $\text{GCN}_{\text{SR}}$ & $\textcolor{redtext}{69.4 {\tiny\pm1.5}}^{+}$ & $72.0 {\tiny\pm1.0}$ & $24.4 {\tiny\pm7.0}^{-}$ & $73.5 {\tiny\pm1.5}$ & $90.3 {\tiny\pm0.6}^{-}$ & $74.7 {\tiny\pm3.2}^{-}$ \\
 \hline
 \multicolumn{7}{l}{\textit{Multi}} \\
 $\text{GAT}_{\text{MR}}$ & $\textcolor{orangetext}{69.0 {\tiny\pm1.7}}^{+}$ & $\textcolor{redtext}{75.5 {\tiny\pm1.2}}^{+}$ & $29.9 {\tiny\pm4.5}$ & $73.6 {\tiny\pm0.8}^{+}$ & $\textcolor{orangetext}{96.2 {\tiny\pm0.4}}$ & $89.7  {\tiny\pm1.0}^{-}$ \\
 $\text{GCN}_{\text{MR}}$ & $\textcolor{orangetext}{69.0 {\tiny\pm2.5}}^{+}$ & $\textcolor{orangetext}{75.2 {\tiny\pm1.3}}^{+}$ & $\textcolor{redtext}{38.6 {\tiny\pm4.6}}$ & $\textcolor{redtext}{76.1 {\tiny\pm1.2}}^{+}$ & $\textcolor{redtext}{96.3 {\tiny\pm0.2}}$ & $91.9  {\tiny\pm0.3}^{-}$ \\
 \hline
\end{tabular}
\caption{Test accuracy of the models, categorized by encoder type (GAT or GCN), structure (single encoder S or multi-encoder M), and the type of graphs each encoder had access to—either only the original/base representation (B) or also the renormalized representations (R). The highest average values for each dataset are highlighted in \textcolor{redtext}{red}, while the second-highest averages are highlighted in \textcolor{orangetext}{orange}. Next to the results of the renormalized models, a '+' indicates that they performed significantly better than their corresponding base model at the 5\% significance level, according to the Wilcoxon test. Conversely, a '-' indicates that the base model was significantly better. If no symbols are present, the two models are considered statistically equivalent.}
\label{tab:comparison2}
\end{table*}


Models using multiple graphs with renormalized representations, indicated with the subscript MR, almost always have the highest average value and often the second highest as well. It is also noticeable that the second-highest value is sometimes reached by models that use multiple parallel encoders but process the same graph, indicated with the subscript MB. This could suggest that the model benefits from the larger number of parameters. Additionally, MR models not only provide extra parameters that seem useful for the task but also, by processing graphs with different representations, are less prone to overfitting and learn the task more effectively.

Looking more closely at the results, particularly at the statistical tests, it appears that using only the renormalized representation (subscript SR) can be beneficial or not, depending on the case, compared to using a single encoder that processes only the original graph. In contrast, multi-encoder models where the first encoder processes the original graph while the others process the renormalized graph at characteristic scales are almost always statistically equivalent or superior to analogous models that only see the original graph.

One interpretation is that the renormalized graph representation encodes different information than what is accessible to a network processing only the original graph, explaining the statistically significant advantage of MR models. However, this scale is not always the most effective for solving the task, leading to the fluctuating performance of SR models.

Finally, it should be noted that the renormalized representation in the \textit{Computers} dataset negatively impacts performance. This could be because the original graph is already in an optimal representation for the task, but further studies should rigorously investigate this question. Additionally, this dataset has the highest number of edges, suggesting that the representation’s limitations might be related to graph size in these terms.

\begin{figure}[ht!] 
    \centering
    \begin{subfigure}[b]{\linewidth}
        \centering
        \includegraphics[width=\linewidth]{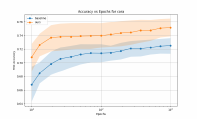}
        \caption{GCN graph encoder}
        \label{fig:graph1}
    \end{subfigure}
    \hfill
    \begin{subfigure}[b]{\linewidth}
        \centering
        \includegraphics[width=\linewidth]{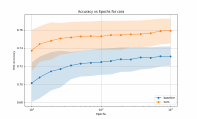}
        \caption{GAT graph encoder}
        \label{fig:graph2}
    \end{subfigure}
    
    

    \caption{Test accuracy of the multiscale model with acces to the renormalized representations, "ours", in orange and its baseline that access only the original graph, "baseline", in blue over epochs for the Cora Dataset.}
    \label{fig:test_acc_vs_epochs}
\end{figure}

In Figure \ref{fig:test_acc_vs_epochs}, we also present the test accuracy along with its confidence intervals as the number of training epochs varies. We present the graphs related to the Cora dataset as an example. The other datasets are available in the supplementary materials.

The fact that the multiscale model consistently outperforms the single-scale model shows that having access to multiple scales introduces a helpful bias for the task during all epochs in the training phase. \\

Having found experimental evidence of the usefulness of multiple graph scales for performing node classification tasks on some graphs, we now turn our attention to the second research question \textbf{RQ2}: can we systematically identify the optimal scales using spectral entropy? To answer this question, we generated graphs aligned with random scales within three ranges: [0, 1), [0, 10), and [0, 100). We then evaluated the statistical significance of these models using a Wilcoxon test on the scores. The alternative hypothesis in this test was that the models found randomly perform better than the multiscale model, which uses the characteristic scale identified by the spectral entropy behavior.

\begin{figure}[ht!] 
    \centering
    \begin{subfigure}[b]{\linewidth}
        \centering
        \includegraphics[width=\linewidth]{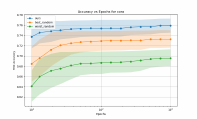}
        \caption{Random interval [0, 100)}
        \label{fig:graph1}
    \end{subfigure}
    \begin{subfigure}[b]{\linewidth}
        \centering
        \includegraphics[width=\linewidth]{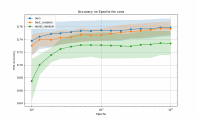}
        \caption{Random interval [0, 10)}
        \label{fig:graph2}
    \end{subfigure}
    
    
    \begin{subfigure}[b]{\linewidth}
        \centering
        \includegraphics[width=\linewidth]{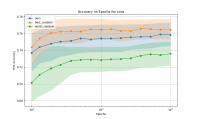}
        \caption{Random interval [0, 1)}
        \label{fig:graph3}
    \end{subfigure}

    \caption{Test accuracy of the model using the characteristic scale ('ours') in blue, the best randomly found scale ('best\_random') in orange, and the worst randomly found scale ('worst\_random') in green across the three intervals, varying by epochs. The values refer to the Cora dataset.}
    \label{fig:random_control}
\end{figure}

Figure \ref{fig:random_control} shows the model using the characteristic scale and the best and worst models obtained with random scales in the three considered ranges for the Cora dataset. Similar graphs for other datasets are presented in the supplementary materials. For each of the three ranges, 10 random values were extracted. These values were used as scales to preprocess the graphs, aligning them to those scales. The random models were compared with the model using the characteristic scale using a Wilcoxon test, with the alternative hypothesis being that one of the random models performs better than the one using the characteristic scale. Due to the presence of multiple comparisons (30 in total), the Bonferroni correction was applied by dividing the significance threshold by the number of comparisons. None of the random scales proved significantly better than the characteristic scale and some are evidently worse performing. 

The best random models identified still had scales close to the characteristic scale; for example, in the Figure \ref{fig:graph3}, the best random model used a scale of $\tau=0.53$ compared to the characteristic scale of $\tau^{\star}=0.57$. It is important to note that the characteristic scale was not found using any form of parameter tuning or trainable hyperparameters. Instead, it was identified a priori by observing the behavior of the spectral entropy of the graph. Therefore, it is a principled value linked to a well-established theory, namely the LRG theory. This situation is quite different from typical parameters learned from data, about which little can be said due to weak initial assumptions that render any interpretation vague. In contrast, in our case, we know that the characteristic scale is associated with mesoscale structures, identified by nodes that quickly connect in the early stages of an information diffusion process on the graph.

It is also important to note that the advantage of using a different scale does not solely depend on connecting distant nodes. This is evidenced by the fact that using larger scales than the characteristic scale does not yield better results and may, in fact, result in statistically worse performance. This is observed when comparing with models that use random scales over large intervals, where many sampled values were much larger than the characteristic scales of the graphs. Nevertheless, the characteristic scale proved to be the most effective.

Overall, we can answer the research question \textbf{RQ2} positively: the optimal scale can be found a priori in a principled manner by observing the behavior of spectral entropy. This is an unexpected and interesting finding of the study, as it allows, a priori—before any training—to define the preprocessing experimentally identified as the best in cases where the renormalized representation has proven to be useful.\\

Experimental results have shown that, in the task of node classification, providing the GNN with information about the mesoscopic structures of the graph can be beneficial. This advantage was observed across all datasets except one, which is also the largest in terms of the number of edges. In cases where observing multiple scales of the graph is useful, the results indicated that the optimal scale is the one found a priori by examining the behavior of spectral entropy, referred to as the characteristic scale. Other scales yielded worse or equivalent results. Furthermore, alternative scales would need to be determined through fine-tuning or heuristics and would lack theoretical interpretations.

It’s not trivial that knowing the graph’s structure at multiple scales would be useful for node classification tasks. It’s also important to note that giving the final classifier only the encoded coarse-scale graphs, without the encoded original graph, worsens performance in some cases. Therefore, both mesoscopic structure information—understanding the node’s relation to these larger-scale structures—and local structure information are necessary.

In light of the positive experimental results and their interpretation, we will now discuss some limitations of this procedure, we then summarize the history of the renormalization group, which is well known in physics but less familiar in other fields, and ultimately draw conclusions about the usefulness of RG theory in node classification tasks with GNNs.


\section{Limitations}
Here are some limitations of the work that can serve as inspiration for future research directions. 

The procedure we presented involves identifying characteristic scales and rewiring the graphs to align them with these scales, subsequently providing a model with graphs aligned at different scales. LRG theory applies to graphs with a single connected component and undirected edges. The first limitation is not too significant since we can always treat a graph with multiple connected components as separate graphs. The second limitation, being undirected, is shared by many GNN architectures, and it's possible this could be overcome in the future. Similarly, we didn't consider weighted graphs or edge features, which would require deviating further from the LRG theory, which focuses solely on the graph topology.

For the same reasons, spectral entropy does not account for node features or the specific task. An interesting direction for future work is to define spectral entropies that incorporate features and possibly task-specific information. This will be one of our next steps, along with applying these techniques to graph classification tasks, where this method might provide a useful bias, similar to what \cite{ying2018hierarchical} found.

Identifying the characteristic scale and performing preprocessing can be computationally intensive, but it's a process that only needs to be done once. Despite graph sparsification, preprocessing might sometimes increase the number of edges compared to the original graph, and handling multiple graphs could require more memory. However, each scale can be processed in parallel and understanding how to efficiently analyze multiple scales is another research direction that we find highly interesting. An analysis of algorithm execution times and the number of edges in the preprocessed graphs is provided in the supplementary materials.

\section{Renormalization Group}
We briefly summarize some of the key milestones in the history of the Renormalization Group (RG), while also drawing connections to statistical physics of networks and graph learning. 

RG theory is a powerful framework describing the change in the mathematical representation of a system when looked at different scales. 
RG was first introduced in quantum electrodynamics \cite{bethe1947electromagnetic} to remove the infinities that arise from the small-scale description of the system (we refer the reader to the surveys \cite{lepage1989renormalization} and \cite{delamotte2004hint}). After its first appearance in quantum electrodynamics, the RG reached full maturity with the work on continuous phase transitions by Wilson \cite{wilson1974renormalization} (we refer the reader to the survey \cite{fisher1974renormalization}). Wilson's approach was that of eliminating microscopic degrees of freedom. This was paralleled by Kadanoff's intuition \cite{kadanoff1966scaling} that the strong correlation acting in the critical regime could allow for describing the system using blocks of the initial smaller components. Despite the challenges posed by their strong topological heterogeneity, decades later a renormalization group for complex networks was ultimately found in the Laplacian Renormalization Group (LRG) \cite{villegas2023laplacian, villegas2022laplacian}. Such an approach was able to overcome the limitations of previous attempts \cite{garcia2018multiscale} like the impossibility to reconnect it to ordinary renormalization when applied to regular lattices. To our knowledge, the present paper is the first attempt to use LRG in combination with GNNs. Nonetheless, since RG is deeply connected with diffusion processes, it's important to remember that diffusion processes have been used as a pre-pocessing step in graph learning \cite{gasteiger2018predict,gasteiger2019diffusion}.

\section{Conclusions}
Graphs are complex structures with a wealth of information embedded in their topology. In particular, a graph provides information about explicit and local interactions through the neighborhood of each node, but it also encodes implicit, long-range interactions between nodes connected by a path. In principle, GNN layers, when iterated multiple times, can capture both types of information. However, it is well known that the limited space of representations restricts the range over which information can be transmitted \cite{alon2020bottleneck}. The best solution to this problem is to find new graph representations that encode long-range interactions \cite{di2023over}.

In this work, we present for the first time representations based on a renormalization group theory for graphs. Our goal is to help bridge the gap between network physics and graph learning—two fields that, despite the potential for fruitful idea exchange, remain largely separate and difficult to reconcile in terms of terminology and concepts. Specifically, we establish a possible link between the Laplacian Renormalization Group and graph rewiring for long-range information transmission.

The fact that in our experiments models with access to both the original and renormalized graph representations were statistically equivalent to or better than those processing only the original graph supports the hypothesis that the renormalized representation provides GNN layers with different information than the original representation. The fact that this is only sometimes beneficial likely depends on the task. Given a graph and the various types of information encoded in its topology, not all of these may be necessary or useful for the specific task the model is evaluated on.

For the \textit{Computers} dataset in particular, future studies should investigate whether the disadvantage associated with the renormalized representation is task-related or due to the dataset’s size, as it was the largest among those analyzed. Interestingly, many of the datasets where the renormalized representation was beneficial are citation networks, which generally have a hierarchical, multi-scale structure \cite{ravasz2003hierarchical}. This observation suggests promising directions for future research.

Among the datasets where the renormalized representation was useful, experiments showed that the characteristic scale—defined as the point in the diffusion process where rapid homogenization occurs within certain mesostructures before a slower phase leading to full homogenization—was also the optimal one experimentally. No other randomly extracted scale performed statistically better, and many performed worse.

The fact that, for certain tasks and datasets, this hyperparameter can be determined a priori based on theoretical reasoning on the spectral entropy—before analyzing any task-related data and relying solely on graph topology—is noteworthy. Further investigation into its connections with graph learning is warranted for future research.


Both research questions addressed in this work suggest that RG theory can offer a solid foundation for deep learning, and specifically, that LRG theory can do so for graph deep learning. The multi-scale approach offered by LRG holds promise for improving node classification, as demonstrated in this work. We hope that future research will confirm its potential in other tasks like link prediction and graph classification.





\section{Acknowledgements}
This work was partially supported by projects FAIR (PE0000013) and SERICS (PE00000014) under the MUR National Recovery and Resilience Plan funded by the European Union - NextGenerationEU. Supported also by the ERC Advanced Grant 788893 AMDROMA,  EC H2020RIA project “SoBigData++” (871042), PNRR MUR project  IR0000013-SoBigData.it.


\end{document}